\documentclass[runningheads]{llncs}
\usepackage{graphicx}
\usepackage{graphicx}

\usepackage{booktabs}       
\usepackage{amsfonts}       
\usepackage{nicefrac}       
\usepackage{microtype}      
\usepackage{xcolor}         
\usepackage{mathrsfs}  
\usepackage{bm}
\usepackage{amssymb,amsmath}

\usepackage{subcaption}
\usepackage{wrapfig}
\usepackage{picinpar}
\usepackage{cutwin}
\usepackage{graphicx,multicol}
\usepackage{algorithm}  
\usepackage{algorithmic}
\usepackage{stmaryrd}

\usepackage{multirow}
\usepackage{algorithm}  
\usepackage{algorithmic}
\usepackage{amsfonts}
\usepackage{amsfonts,amssymb} 
\usepackage{verbatim}
\usepackage{url}
\usepackage{enumitem}
\usepackage{hyperref}
\begin{document}
\title{Clinical Domain Classification from Medical Transcriptions}

%
\titlerunning{ Clinical Domain Classification}

%
%


\author{Sravani Pottipati\inst{1} \and Lakshmikar R. Polamreddy\inst{2}}
\authorrunning{Pottipati et al.}
\institute{Independent Researcher, New York City, USA,\\
\email{sravlaks18112015@gmail.com}
\and
Yeshiva University, New York City, USA,\\
\email{lpolamre@mail.yu.edu}
}
\maketitle              

\begin{abstract}
Clinical domain classification plays an important role in organizing and analyzing large volumes of unstructured medical text. However, medical transcription datasets are often highly imbalanced, which can substantially degrade classification performance, particularly for underrepresented clinical specialties. In this work, we present a comparative study of machine learning and transformer-based approaches for clinical domain classification from medical transcriptions. We evaluate six traditional machine learning classifiers---Naive Bayes, Support Vector Machine (SVM), Decision Tree, Random Forest, K-Nearest Neighbors (KNN), and XGBoost---along with two pretrained transformer models, BERT and XLNet, and a few-shot large language model prompting approach. Experiments are conducted on medical transcription data collected from MTSamples, comprising 5,013 samples across 40 clinical specialties. To address severe class imbalance, we investigate two balancing strategies: text augmentation using NLP-based synonym replacement and Synthetic Minority Over-sampling Technique (SMOTE). Experimental results demonstrate that data balancing substantially improves classification performance across the evaluated models. In particular, BERT achieves the highest F1-score of 0.996 on the SMOTE-balanced dataset while requiring lower training time than XLNet. The results highlight the effectiveness of transformer-based representations combined with appropriate data balancing strategies for clinical domain classification and provide a systematic comparison of classical machine learning, transformer models, and few-shot prompting for medical transcription analysis.

\keywords{Clinical domain classification  \and BERT \and XLNet \and NLP Augmenter \and SMOTE.}
\end{abstract}
\section{Introduction}

In recent years, the volume of clinical data created by electronic health records (EHRs), medical literature, and patient-provided data has increased rapidly in the healthcare business. Efficient medical record classification helps with patient diagnosis, treatment recommendations, disease surveillance, and resource allocation. It is critical for effective patient care, medical research, and decision-making to extract valuable insights from this massive volume of unstructured clinical information. Furthermore, by facilitating data-driven studies, it aids evidence-based medicine, improves healthcare workflow, and helps medical research. Clinical domain classification is dividing medical text data into several classes based on the domain or medical speciality, which is important for organizing and evaluating healthcare data.

AI and NLP have emerged as powerful tools for processing and understanding clinical text.NLP algorithms can interpret unstructured medical language, including notes from doctors, findings from radiology and pathology tests, and summaries of patient discharges. They make it possible to retrieve pertinent data from a variety of clinical papers, including symptoms, diagnosis, therapies, and patient demographics. AI models have performed remarkably well in a variety of NLP tasks, including clinical domain categorization, especially transformer-based models like BERT and XLNet.

This study's main goal is to compare and assess how well various NLP algorithms perform in classifying clinical domains. To test if they are appropriate for managing unbalanced clinical text data, we evaluate both conventional machine learning classifiers and transformer-based models. We will specifically look into how data balancing approaches affect model performance. we explore various state-of-the-art NLP algorithms for clinical domain classification. We aim to evaluate their performance on both balanced and unbalanced datasets to understand their robustness in handling imbalanced clinical text data.

The remainder of the paper is organized as follows. Section~\ref{sec:rel} covers seminal contributions made in the filed of NLP applications in healthcare. Section~\ref{sec:methods} explains the models and their architectures used for clinical domain classification. Section~\ref{sec:data} presents the dataset and Section~\ref{sec:results} compares performance of all the models and their results. Section~\ref{sec:chal} discusses challenges we face during the course of this work. Section~\ref{sec:concl} captures concluding remarks of the proposed work and scope for future work.

\section{Related Work}\label{sec:rel}

Varshini et al.~\cite{varshini2023extraction} proposed a web scraping method to extract the clinical notes from the Medical Transcription (MT) samples which hold many transcripted clinical notes of various departments. In addition, Natural Language Processing (NLP) was used to pre-process the data, and the variants of the Term Frequency-Inverse Document Frequency (TF-IDF)-based vector model are used for the feature selection, thus extracting the required data from the clinical notes. The performance measures including the accuracy, precision, recall and F1 score are used in the identification of disease. They concluded that the result obtained proveed that the Random Forest Classifier obtained a higher accuracy of 90 percent when compared to the other algorithms.

Li et al.~\cite{li2019joint} proposed a comprehensive model based on CMed-BERT, RCNN and BiGRU-CRF for a joint task of department identification and slot filling of the specific domain. Experimental results confirmed the competitiveness of their model though the medical vocabulary and clinical entities in different departments of the hospital often differ to some extent.

Yao et al.~\cite{yao2019traditional} proposed the problem of classifying TCM(Traditional Chinese Medicine) clinical records into 5 main disease categories. They explored a number of state-of-the-art deep learning models and found that the recent Bidirectional Encoder Representations from Transformers can achieve better results than other deep learning models and other state-of-the-art methods.

Lavanya et al.~\cite{lavanya2021deep} proposed to investigate the models of deep learning (DL) techniques applied to classify the text in social media healthcare networks. Their study provided an insight for training the data and to classify the text by analyzing and extracting the raw input and produce the output with the help of Natural language processing (NLP).

Zhou et al.~\cite{zhou2022natural} proposed to elaborate on different NLP approaches and the NLP pipeline for smart healthcare from the technical point of view. Then, they introduced representative smart healthcare scenarios, including clinical practice, hospital management, personal care, public health, and drug development. They further discussed two specific medical issues, i.e., the coronavirus disease 2019 (COVID-19) pandemic and mental health, in which NLP-driven smart healthcare plays an important role.

Wang et al.~\cite{wang2023prompt} proposed a overview of latest advances in prompt engineering in the field of natural language processing (NLP) for the medical domain and emphasized significant contributions to healthcare NLP applications such as question-answering systems, text summarization, and machine translation.

Lamichhane~\cite{lamichhane2023evaluation} proposed to report the performance of LLM-based ChatGPT (with gpt-3.5-turbo backend) in three text-based mental health classification tasks: stress detection (2-class classification), depression detection (2-class classification), and suicidality detection (5-class classification). They concluded that the zero-shot classification accuracy obtained with ChatGPT indicates a potential use of language models for mental health classification tasks.

Vaira et al.~\cite{vaira2018mamabot} focused on exploiting AI-based chatbot systems, mainly based on machine learning algorithms and Natural Language Processing, to understand and respond to needs of patients and their families. In particular, they described an application scenario for an AI-chatbot delivering support to pregnant women, mothers, and families with young children, by giving them help and instructions in relevant situations.

Carchiolo et al.~\cite{carchiolo2019medical} presented concerning the digitization of medical prescriptions, both to provide authorization for healthcare services or to grant reimbursement for medical expenses. The proposed system first extracted text from scanned medical prescription, then Natural Language Processing and machine learning techniques provided effective classification exploiting embedded terms and categories about patient/doctor personal data, symptoms, pathology, diagnosis and suggested treatments.

Zhang et al.~\cite{zhang2021novel} proposed a novel method that combines knowledge-graph-based and word-embedding-based similarity measures via word entropy. An experiment is conducted on five public datasets (R$\&$G, M$\&$C, WS353, WS353-Sim and SimLex). The experimental results show that the proposed method achieves significant improvements over other word similarity measures in terms of the correlation coefficient.

\section{Methods}\label{sec:methods}
\begin{figure}
\begin{center}
\includegraphics[width=1.35\textwidth]{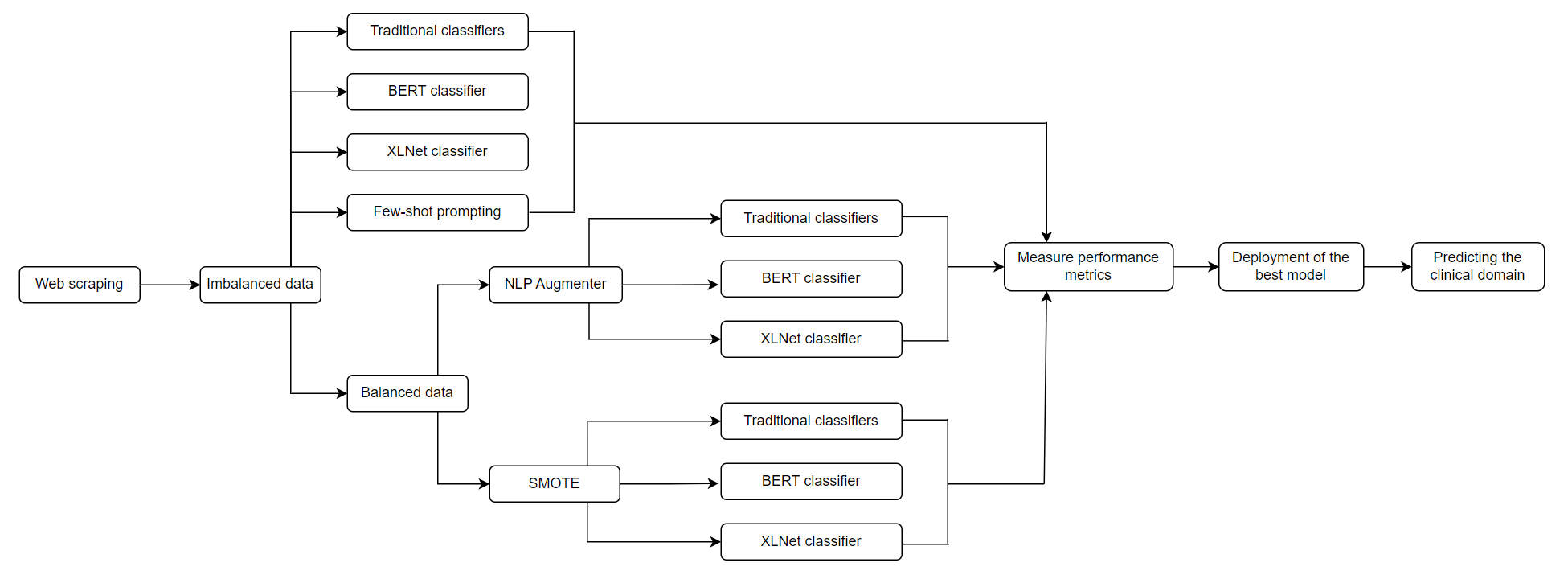}
\caption{Model building approach} \label{fig:appr}
\vspace{-0.6cm}
\end{center}
\end{figure}

We handle this clinical classification task as per the approach mentioned in the fig.~\ref{fig:appr}. We attempt to explore the various models like traditional classifiers ( SVM, KNN, Decision-tree, Random forest, Naive Bayes, XGBoost), pre-trained transformer models (BERT, XLNet) and Few-shot prompting. First, we test the performance of all these models with the original dataset obtained from MTsamples website. Later, we convert the original imbalanced dataset into balanced dataset using NLP Augmenter and then measure the performance of these models. In addition, we also apply SMOTE algorithm to create balanced dataset and then performance is measured. Among all these models, we select the best model in terms of F1 score and computational cost to deploy and then to use for prediction of the clinical domain.

\subsection{Traditional classifier models}\label{sec:trad}
We use these traditional classifier models for this task of classification - Naive Bayes is a probabilistic classification algorithm known for its simplicity and efficiency. SVM (SUpport Vector Machine) is a powerful classification algorithm that aims to find the best hyperplane to separate data points into different classes. Decision Trees are non-linear classifiers that create a tree-like model for making decisions based on features' values. Random Forest is an ensemble method that combines multiple Decision Trees to improve prediction accuracy. K-Nearest Neighbors (KNN) is a simple yet effective algorithm that classifies data points based on the majority class among their nearest neighbors. XGBoost is a gradient boosting algorithm known for its high accuracy and efficiency.

\subsection{Transformer models}\label{sec:trans}
\begin{figure}
\begin{center}
\includegraphics[width=0.9\textwidth]{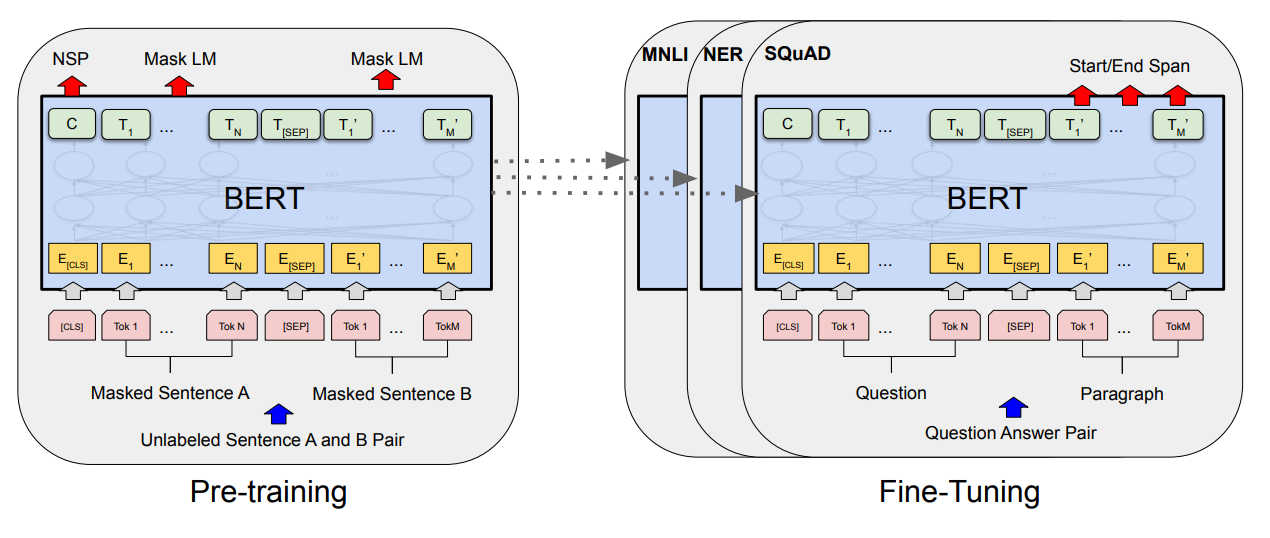}
\caption{Overall pre-training and fine-tuning procedures for BERT~\cite{devlin2018bert}} \label{fig:BERT}
\vspace{-0.8cm}
\end{center}
\end{figure}

We make use of pretrained BERT model, which stands for Bidirectional Encoder Representations from Transformers. BERT’s model architecture~\cite{devlin2018bert} is a multi-layer bidirectional Transformer encoder based on the original implementation described in Vaswani et al.~\cite{vaswani2017attention} in 2017 and released in the tensor2tensor library. The overall pre-training and fine-tuning procedures for BERT is shown in fig.~\ref{fig:BERT}.

In addition to BERT, we also utilize XLNet model~\cite{yang2019xlnet},  a generalized autoregressive pretraining method for clinical text classification. XLNet integrates ideas from Transformer-XL, the state-of-the-art autoregressive model, into pre-training.

\subsubsection{Implementation details}-
Both the BERT and XLNet models are trained using the following hyperparameters:
\begin{enumerate}
    \item Number of epochs: 20
    \item Batch size: 16
    \item Optimizer: AdamW
    \item Learning rate: 2e-5
\end{enumerate}

\subsection{Few-shot prompting}\label{sec:fewshot}
In few-shot prompting technique, we tune an LLM (Large Language Model) - "gpt-3.5-turbo" using instructions. In the instructions, we include two examples for just two out out of forty categories and then the model is used to predict the clinical domain based on the medical transcription. Iterative prompting that would lead to better prompts could improve the results. The main advantages of using an instruction based LLMs are better performance with no training and very less computational cost.

\section{Datasets}\label{sec:data}
We use the data from MTSamples.com website  that offers access to a large collection of transcribed digital medical reports and examples. The site is designed to cater to both learning and working medical transcriptionists, providing them with sample reports for various medical specialties and work types. These samples are contributed by various transcriptionists and users and are intended for reference purposes only.
This website offers 5013 samples across 40 different medical specialties as shown in the table~\ref{tab:med}. We perform web scraping to extract this data into a .csv file from this website. This data obtained from the website is highly imbalanced with 'Surgery' category holding 1105 samples and many categories holding just tens of samples.

\begin{table}
\begin{center}
\caption{Medical transcription data provided by MTSamples.com}\label{tab:med}
\setlength{\tabcolsep}{+0.6cm}{
\begin{tabular}{lcc}
\hline
Medical specialty & Number of samples\\
\hline
Allergy / Immunology  & 8 \\
Autopsy & 8 \\
Bariatrics & 18 \\
Cardiovascular / Pulmonary  & 372 \\
Chiropractic  & 14 \\
Consult - History and Phy. & 515 \\
Cosmetic / Plastic Surgery & 27 \\
Dentistry & 27 \\
Dermatology & 30 \\
Diets and Nutritions & 10 \\
Discharge Summary & 108 \\
Emergency Room Reports  & 75 \\
Endocrinology & 19 \\
ENT - Otolaryngology & 99 \\
Gastroenterology & 230 \\
General Medicine & 259 \\
Hematology - Oncology & 90 \\
Hospice - Palliative Care & 6 \\
IME-QME-Work Comp etc. & 16 \\
Lab Medicine - Pathology & 8 \\
Letters & 24 \\
Nephrology & 81 \\
Neurology & 223 \\
Neurosurgery & 94 \\
Obstetrics / Gynecology & 160 \\
Office Notes & 53 \\
Ophthalmology & 83 \\
Orthopedic & 359 \\
Pain Management & 63 \\
Pediatrics - Neonatal & 70 \\
Physical Medicine - Rehab & 21 \\
Podiatry & 48 \\
Psychiatry / Psychology & 53 \\
Radiology & 273 \\
Rheumatology & 10 \\
Sleep Medicine & 20 \\
SOAP / Chart / Progress Notes & 167 \\
Speech - Language  & 9 \\
Surgery & 1105 \\
Urology & 158 \\
\hline    
\end{tabular}}
\end{center}
\end{table}

Imbalanced data can lead to biased model performance, where the model may perform well on the majority class but poorly on the minority class. When we train all the models with this data, we notice that the models under-performed substantially in terms of overall accuracy and F1 score. So, we use two different approaches to generate synthetic data to handle data imbalance issue. One is text augmentation using NLP Augmenter library, using which we generate 1500 samples for each category. This augmenter is used to perform text augmentation by replacing words with their synonyms. The other approach is using SMOTE( Synthetic Minority Over-sampling Technique) algorithm. This algorithm creates synthetic examples by interpolating between existing minority samples. It selects a minority sample and its k-nearest neighbors, and then it generates new samples along the line segments joining the selected sample and each of its neighbors. After applying SMOTE, the minority class is augmented with synthetic samples, creating a balanced dataset where all classes have approximately the same number of samples. We use these two imbalanced datasets to train and measure the performance metrics of the models.

\section{Results}\label{sec:results}
In this paper, we evaluate several classical machine learning models along with two transformer-based models, BERT and XLNet and few-shot prompting technique with an LLM. The evaluation metrics used to compare the models are accuracy, precision, recall,F1-score, and training time. The results otained using the original imbalanced dataset is shown in the table~\ref{tab:imb}. We notice that all the models show poor performance and overall F1-score could not go beyond 0.20 for nay model due to imbalanced nature of the data. In terms of computational cost, KNN and Naive Bayes demonstrated better results. Few-shot prompting technique with an LLM using an API call could produce highest accuracy with 0.37. Data imbalance needs to be addressed to improve performance of the models.

\begin{table}
\begin{center}
\caption{Performance metrics of the models with original imbalanced data}\label{tab:imb}
\setlength{\tabcolsep}{+0.1cm}{
\begin{tabular}{lcccccc}
\hline
Model & Training time(in secs) & Accuracy & Precision & Recall & F1-score\\
\hline
Naive Bayes & \textbf{0.09} & 0.32 & 0.16 & 0.32 & 0.19 \\
SVM & 87.88 & 0.19 & 0.17 & 0.19 & 0.17 \\
Decision tree & 4.44 & 0.06 & 0.06 & 0.06 & 0.06 \\
Random forest & 141.72 & 0.08 & 0.08 & 0.08 & 0.08 \\
KNN & \textbf{0.06} & 0.21 & 0.20 & 0.21 & \textbf{0.20} \\
XGBoost & 1324.33 & 0.09 & 0.09 & 0.09 & 0.09 \\
XLNet & 710.88 & 0.22 & 0.19 & 0.22 & \textbf{0.20} \\
BERT & 599.94 & 0.22 & 0.21 & 0.22 & \textbf{0.20} \\
Fewshot prompting & - & \textbf{0.37} & - & - & - \\
\hline    
\end{tabular}}
\end{center}
\end{table}

The results obtained using the balanced data generated by NLP augementer are shown in table~\ref{tab:aug}. We notice that SVM outperformed all other models in terms of F1-score with 0.77 followed by XLNet model with 0.76 but the training time of these models are substantially higher when compared to the other models. KNN model takes just 0.59 sec for training and shows moderately good F1-score with 0.74.

\begin{table}
\begin{center}
\caption{Performance metrics of the models with imbalanced data generated by NLP Augmenter}\label{tab:aug}
\setlength{\tabcolsep}{+0.1cm}{
\begin{tabular}{lcccccc}
\hline
Model & Training time(in secs) & Accuracy & Precision & Recall & F1-score\\
\hline
Naive Bayes & \textbf{1.05} & 0.75 & 0.72 & 0.75 & 0.72 \\
SVM & 9721.08 & 0.79 & 0.76 & 0.79 & \textbf{0.77} \\
Decision tree & 491.62 & 0.70 & 0.70 & 0.70 & 0.70 \\
Random forest & 604.32 & 0.76 & 0.73 & 0.76 & 0.74 \\
KNN & \textbf{0.59} & 0.76 & 0.73 & 0.76 & 0.74 \\
XGBoost & 3702.52 & 0.75 & 0.73 & 0.75 & 0.74 \\
XLNet & 14154.93 & 0.78 & 0.75 & 0.78 & \textbf{0.76} \\
BERT & 11880 & 0.76 & 0.74 & 0.76 & 0.74 \\
\hline    
\end{tabular}}
\end{center}
\end{table}

\begin{figure}
\begin{center}
\includegraphics[width=0.9\textwidth]{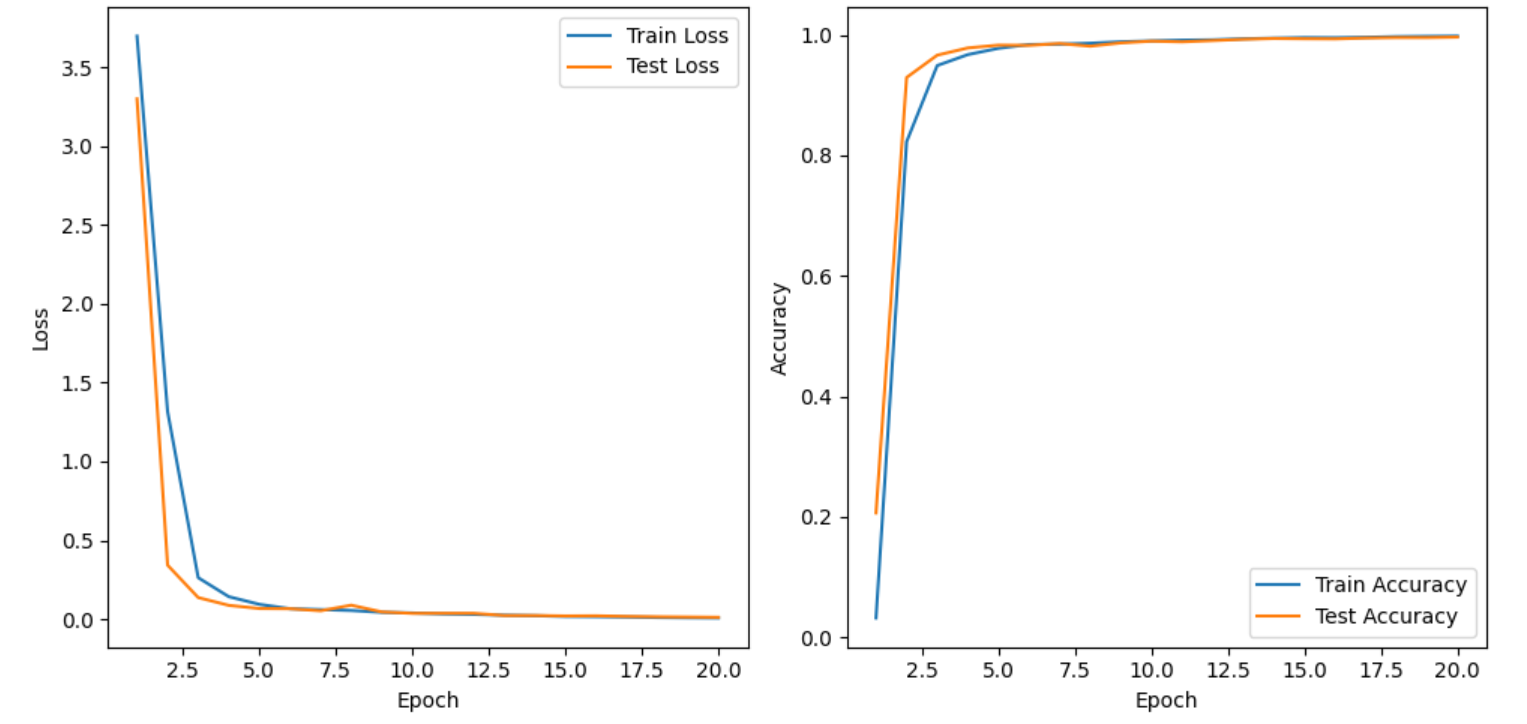}
\caption{BERT model accuracy and loss versus epochs} \label{fig:acc}
\vspace{-0.8cm}
\end{center}
\end{figure}

The results obtained using the balanced data generated by SMOTE algorithm are shown in table~\ref{tab:smote}. We notice that BERT and XLNet models outperformed all other models in terms of F1-score with 0.99 and BERT is computationally less expensive when compared to XLNet. In this case also, KNN model takes just 0.29 sec for training and but its F1-score stands at 0.81. The BERT results for accuracy and loss versus epoch are shown in fig.~\ref{fig:acc}. We run BERT model for just 20 epochs as the accuracy dramatically improves and loss converges from second epoch.

\begin{table}
\begin{center}
\caption{Performance metrics of the models with imbalanced data generated by SMOTE algorithm}\label{tab:smote}
\setlength{\tabcolsep}{+0.1cm}{
\begin{tabular}{lcccccc}
\hline
Model & Training time(in secs) & Accuracy & Precision & Recall & F1-score\\
\hline
Naive Bayes & \textbf{2.07} & 0.82 & 0.79 & 0.82 & 0.80\\
SVM & 1036.42 & 0.83 & 0.81 & 0.83 & 0.81 \\
Decision tree & 131.84 & 0.78 & 0.78 & 0.78 & 0.78 \\
Random forest & 306.14 & 0.81 & 0.81 & 0.82 & 0.81 \\
KNN & \textbf{0.29} & 0.83 & 0.79 & 0.83 & 0.81 \\
XGBoost & 9084.54 & 0.83 & 0.82 & 0.83 & 0.82 \\
XLNet & 7896.69 & 0.99 & 0.99 & 0.99 & \textbf{0.99} \\
BERT & 6172.77 & 0.996 & 0.996 & 0.996 & \textbf{0.996} \\
\hline    
\end{tabular}}
\end{center}
\end{table}

From these results, we observe that all the models show significantly better performance with balanced data when compared to the original imbalanced data. If we compare the results between the balanced data of NLP Augmenter and SMOTE, all the models exhibit better performance when trained on the data of SMOTE, especially transformer models. It's worth noting that transformer models are known to be capable of capturing contextual information and handling data variations well, so the impact of data augmentation using NLP Augmenter might not be as significant for them as SMOTE. In summary, NLP Augmenter plays a crucial role in enhancing the performance of classical machine learning models by providing additional data samples through synonym replacement. However, its impact on transformer-based models might not be as pronounced as SMOTE.

\section{Challenges}\label{sec:chal}
The challenges that we face during this project are listed below.
\begin{enumerate}
    \item Imbalanced data: The imbalanced data available on the MTsamples website is a major challenge for us as the models performance is sub-optimal. Dealing with this imbalance is crucial, as it can lead to biased predictions and poor performance on the minority class. We try to handle this issue by generating synthetic samples using two different approaches - NLP Augmenter and SMOTE algorithm. The use of SMOTE effectively addressed this challenge and improved models performance.
    \item Exploding gradients: When we face this issue, the loss value gradually increases with epochs when training with BERT and XLNet models. We try to handle this issue with gradient clipping and scheduling the learning rate.
    \item Computational resources: Training and predicting with transformer-based models like BERT and XLNet can be computationally intensive, requiring substantial resources and time. This could be a challenge, particularly when working with limited computational power or large datasets. However, BERT is less expensive compared to XLNet.
    \item Overfitting with augmented data: Augmented data can introduce the risk of overfitting, where the model becomes too specific to the training data and fails to generalize to new data. Monitoring and controlling overfitting are essential to ensure the model's robustness.
    \item Hyper-parameter Tuning: Optimizing hyper-parameters for each model is essential to achieve optimal performance. However, this process can be time-consuming and requires careful experimentation to strike the right balance between model complexity and performance. 
\end{enumerate}

\section{Conclusion}\label{sec:concl}
In this project, we first scrape the medical transcription data with associated clinical domains from the MTsamples.com website and then build several models for clinical domain classification task. As we notice sub-optimal performance of all the models due to imbalanced nature of the data, we performed data augmentation using NLP augmenter and SMOTE algorithm. With this balanced data, we again measured the performance of these models. We conclude that BERT model outperformed all other models in terms of F1-score with 0.99 and is computationally less expensive than XLNet. Our future work aims to focus on gathering more real medical transcriptions data instead of generating synthetic samples and then these models will be evaluated to identify the best model for this task of clinical domain classification.


%
%
%
\bibliographystyle{splncs04}
\bibliography{mybibliography}
%




\end{document}